\def\arxivtemplate{0}
\def\newarxivtemplate{1}

\documentclass[10pt,twocolumn,letterpaper]{article}

\if\arxivtemplate0
    \usepackage{wacv}
    
    \makeatletter
    \@namedef{ver@everyshi.sty}{}
    \makeatother
    \usepackage{tikz}
    
\fi

\usepackage{times}
\usepackage{epsfig}
\usepackage{graphicx}
\usepackage{amsmath}
\usepackage{amssymb}
\usepackage{booktabs}
\usepackage{float}
\usepackage{tikz}

\usetikzlibrary{positioning, fit}
\graphicspath{ {./imgs/} }

\if\arxivtemplate0
    \def\wacvPaperID{1119} %
    \wacvalgorithmstrack   %
    \wacvfinalcopy %

    \ifwacvfinal
        \usepackage[breaklinks=true,bookmarks=false]{hyperref}
        \if\newarxivtemplate1
            
            \usepackage[title]{appendix}

            \usepackage{titling}
        \fi
    \else
        \usepackage[pagebackref=true,breaklinks=true,colorlinks,bookmarks=false]{hyperref}
    \fi

\else 
    \usepackage[breaklinks=true,bookmarks=false]{hyperref}

    \usepackage[none]{hyphenat}
    \usepackage[font=small]{caption} 
\fi

\begin{document}
\setcounter{page}{1}

\title{Matching with AffNet based rectifications}

\author{Václav Vávra  \\
VRG, Faculty of Electrical Engineering,\\
Czech Technical University in Prague\\
{\tt\small vavravac@fel.cvut.cz} 
\and
Dmytro Mishkin \\
VRG, Faculty of Electrical Engineering,\\
Czech Technical University in Prague,\\
and HOVER Inc. \\
{\tt\small ducha.aiki@gmail.com} 
\and
Jiří Matas\\  
VRG, Faculty of Electrical Engineering,
Czech Technical University in Prague\\
{\tt\small matas@fel.cvut.cz}\\
\\
}
\if\newarxivtemplate1
    \date{}
\fi
\maketitle
\thispagestyle{empty}

\begin{abstract}
We consider the problem of two-view matching under significant viewpoint changes with view synthesis. We propose two novel methods, minimizing the view synthesis overhead. The first one, named DenseAffNet, uses dense affine shapes estimates from ~\cite{AffNet}, which allows it to partition the image, rectifying each partition with just a single affine map. The second one, named DepthAffNet, combines information from depth maps and affine shapes estimates to produce different sets of rectifying affine maps for different image partitions. DenseAffNet is faster than the state-of-the-art and more accurate on generic scenes. DepthAffNet is on par with the state of the art on scenes containing large planes. The evaluation is performed on 3 public datasets -- EVD Dataset~\cite{MODS}, Strong ViewPoint Changes Dataset~\cite{toft} and IMC Phototourism Dataset~\cite{IMC2020}.
\end{abstract}

\section{Introduction}
\label{sec:intro}
Two-view matching is a basic building block of 3D reconstruction and SLAM.
Off-the-shelf algorithms like COLMAP~\cite{colmap1} (based on SIFT~\cite{SIFT}) are mature enough to serve as "ground-truth generators" for various purposes~\cite{IMC2020,CO3D2021}. However, it relies on dense image coverage of the scene, meaning that camera pose difference between two given view is medium at most. 
Sometimes, however, the dense image coverage is not available, e.g. because of problems with physical access to some location, or nature of the data (e.g. historical photography~\cite{MODS-User2020}).

In such cases one needs to resort to more sophisticated methods. 
Such methods mostly rely on test-time augmentation in form of affine~\cite{ASIFT,MODS} or perspective~\cite{toft} view synthesis. We provide a detailed review of them in Section~\ref{sec:pipeline}, which serves both as literature review and method description.

\subsection{Contributions}

This paper makes the following contributions: 
(1) we propose a generalization of the two-view matching pipeline with a rectification preprocessing, supporting methods like those described in~\cite{toft,CnnAssistedCoverings,MODS}. 
(2) we present two novel methods, which are simpler, faster and sometimes  more accurate than the state-of-the-art as measured by the accuracy in epipolar geometry or homography estimation on common datasets as EVD 
Dataset~\cite{MODS}, Strong ViewPoint Changes Dataset~\cite{toft} and IMC Phototourism Dataset~\cite{IMC2020}.

\section{Rectification Pipeline}
\label{sec:pipeline}
\begin{figure*}[htb]
\centering
\begin{tikzpicture}[
every node/.style={inner sep=2,outer sep=2},
node distance = 0mm and 15mm, auto,
grode/.style={rectangle, draw=black!0, very thin, minimum width=5mm},
rectangularnode/.style={rectangle, draw=black!60, very thick, minimum height=2mm},]

\node[rectangularnode, minimum height=22mm, label={input image}] (input_img) {\includegraphics[height=20mm]{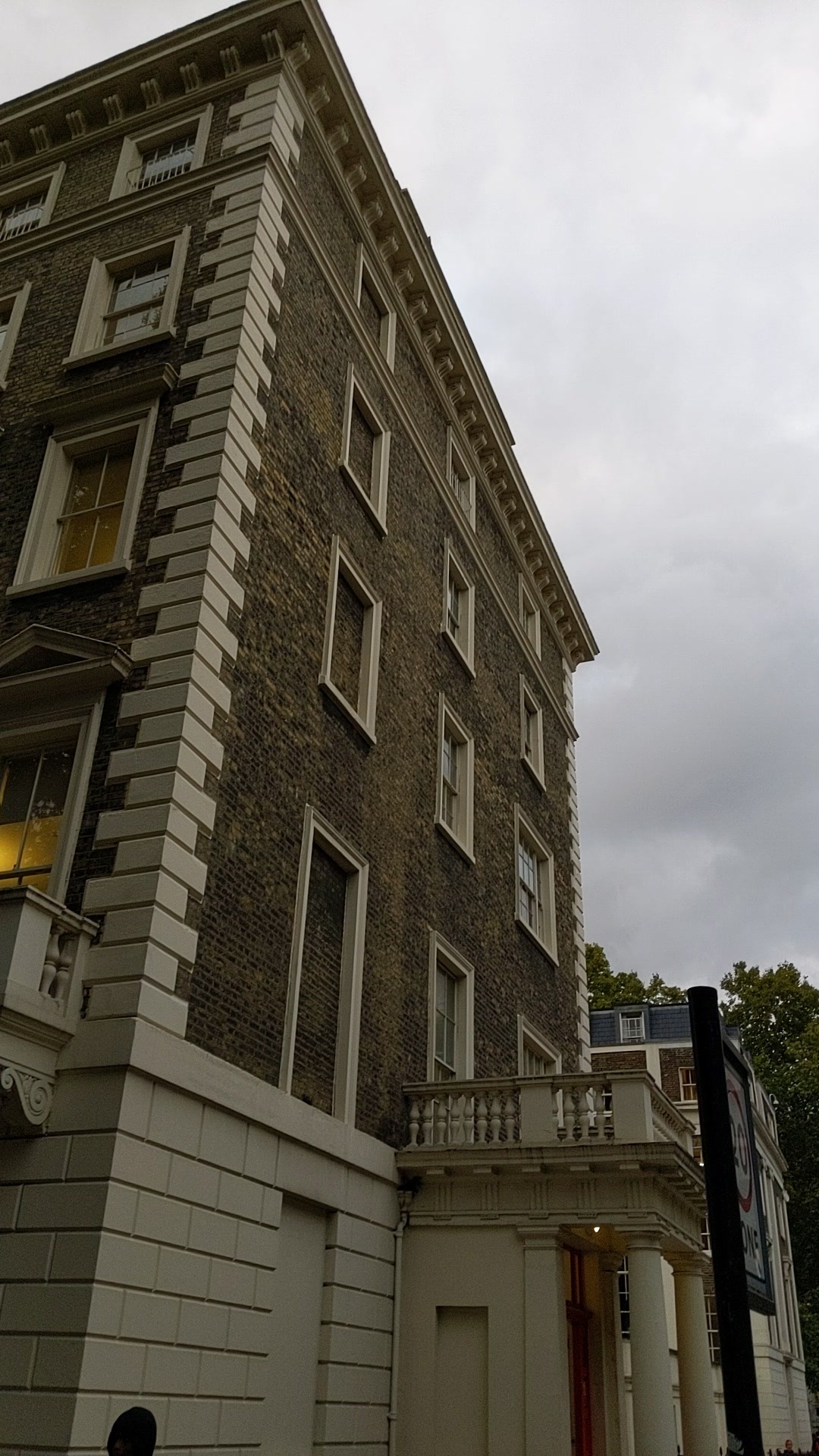}};

\node[rectangularnode, minimum height=22mm, label={segmentation}] (segments) [right=of input_img] {$\{S_s\}_s$};

\node[rectangularnode, label={}] (rect_m_1) [right=of segments, distance=2mm and 2mm] {\{$f_i^{s=2}\}_i$};
\node[rectangularnode, label={[align=center]above:rectifying maps}] (rect_m_2) [above=of rect_m_1, distance=2mm and 2mm] {\{$f_i^{s=1}\}_i$};
\node[rectangularnode, label={}] (rect_m_3) [below=of rect_m_1, distance=2mm and 2mm] {\{$f_i^{s=3}\}_i$};

\node[rectangularnode] (keypoints1) [right=of rect_m_1] {$\{(f_i^{s}(p_{i,j}),D_{i,j,s})\}_{i,j}^{s=2}$};
\node[rectangularnode, label={[align=center]above:keypoint \\ detection and description}] (keypoints2) [right=of rect_m_2] {
$\{(f_i^{s}(p_{i,j}) ,D_{i,j,s})\}_{i,j}^{s=1}$};
\node[rectangularnode] (keypoints3) [right=of rect_m_3] {$\{(f_i^{s}(p_{i,j}),D_{i,j,s})\}_{i,j}^{s=3}$};

\node[rectangularnode, minimum height=22mm, label={[align=center]above:keypoint \\ backprojection}] (backprojection) [right=of keypoints1, align=center] {
$ \{p_{i,j} \}_{i,j,s} $ 
\\
\\
$ \{D_{i,j,s}\}_{i,j,s} $
};

\draw[thick, ->] (input_img.east) -- (segments.west) node[midway,right] {};
\draw[thick, ->] (segments.east|-rect_m_1.west) -- (rect_m_1.west) node[midway,right] {};
\draw[thick, ->] (segments.east|-rect_m_2.west) -- (rect_m_2.west) node[midway,right] {};
\draw[thick, ->] (segments.east|-rect_m_3.west) -- (rect_m_3.west) node[midway,right] {};
\draw[thick, ->] (rect_m_1.east) -- (keypoints1.west) node[midway,right] {};
\draw[thick, ->] (rect_m_2.east) -- (keypoints2.west) node[midway,right] {};
\draw[thick, ->] (rect_m_3.east) -- (keypoints3.west) node[midway,right] {};

\draw[thick, ->] (keypoints1.east) -- (backprojection) node[midway,right] {};
\draw[thick, ->] (keypoints2.east) -- (backprojection) node[midway,right] {};
\draw[thick, ->] (keypoints3.east) -- (backprojection) node[midway,right] {};

\node[grode, label={}] (input) [below=2.1cm of input_img] {\includegraphics[height=22mm]{imgs/input_image.jpg}};

\node[grode, label={}] (input_sgm) [right=8mm of input] {\includegraphics[height=22mm]{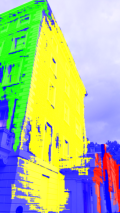}};

\node[grode, label={}] (red) [right=15mm of input_sgm] {\includegraphics[width=.04\textwidth]{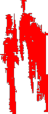}};
\node[grode] (green) [above=0mm of red] {\includegraphics[width=.04\textwidth]{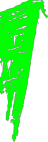}};
\node[grode] (yellow) [below=0mm of red] {\includegraphics[width=.04\textwidth]{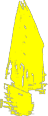}};
\node[grode] (sky) [below=0mm of yellow] {\includegraphics[width=.04\textwidth]{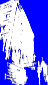}};

\node[grode] (green_top) [below=1.0cm of keypoints1, distance=2mm and 2mm] {\includegraphics[width=.03\textwidth]{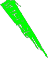}};
\node[grode] (green_first) [below=of green_top,distance=100mm and 10mm] {\includegraphics[width=.03\textwidth]{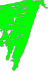}};
\node[grode] (green_third) [below=of green_first,distance=10mm and 100mm] {\includegraphics[width=.03\textwidth]{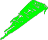}};

\node[grode] (red_right) [below=of green_third,distance=2mm and 2mm] {\includegraphics[width=.03\textwidth]{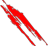}};
\node[grode] (red_first) [below=of red_right,distance=2mm and 2mm] {\includegraphics[width=.03\textwidth]{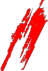}};

\node[grode] (yellow_right) [below=of red_first,distance=2mm and 2mm] {\includegraphics[width=.03\textwidth]{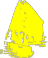}};
\node[grode] (yellow_first) [below=of yellow_right,distance=2mm and 2mm] {\includegraphics[width=.03\textwidth]{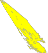}};

\node[grode] (sky_right) [below=of yellow_first, distance=2mm and 2mm] {\includegraphics[width=.03\textwidth]{imgs/frame_out_sky.png}};

\node[grode] (all_back) [right=of red_right, below=2.1cm of backprojection, distance=2mm and 2mm] {\includegraphics[height=22mm]{imgs/frame_out.png}};

\draw[thick, ->] (input) -- (input_sgm) node[midway,above] {};

\draw[thick, ->] (input_sgm) -- (red.west) node[midway,above] {};
\draw[thick, ->] (input_sgm) -- (green.west) node[midway,above] {};
\draw[thick, ->] (input_sgm) -- (yellow.west) node[midway,above] {};
\draw[thick, ->] (input_sgm) -- (sky.west) node[midway,above] {};

\draw[thick, ->] (green) -- (green_top.west) node[midway,above=-0.5mm] {\scriptsize{$f_1^1$}};
\draw[thick, ->] (green) -- (green_first.west) node[midway,above=-0.5mm] {\scriptsize{$f_2^1$}};
\draw[thick, ->] (green) -- (green_third.west) node[midway,above=-0.5mm] {\scriptsize{$f_3^1$}};

\draw[thick, ->] (green_top) -- (all_back) node[pos=0.15,above=1.5mm] {{\scriptsize $(f_1^1)^{-1}$}};
\draw[thick, ->] (green_first) -- (all_back) node[pos=0.15,above=0.5mm] {{\scriptsize$(f_2^1)^{-1}$}};
\draw[thick, ->] (green_third) -- (all_back) node[pos=0.15,above=-0.2mm] {{\scriptsize$(f_3^1)^{-1}$}};

\draw[thick, ->] (red) -- (red_right.west) node[midway,above] {{\scriptsize$f_1^2$}};
\draw[thick, ->] (red) -- (red_first.west) node[midway,above=-1mm] {{\scriptsize$f_2^2$}};

\draw[thick, ->] (red_right) -- (all_back) node[pos=0.15,above=-1mm] {{\scriptsize$(f_1^2)^{-1}$}};
\draw[thick, ->] (red_first) -- (all_back) node[pos=0.15,above=-1mm] {{\scriptsize$(f_2^2)^{-1}$}};

\draw[thick, ->] (yellow) -- (yellow_right.west) node[midway,above] {{\scriptsize$f_1^3$}};
\draw[thick, ->] (yellow) -- (yellow_first.west) node[midway,above=-1mm] {{\scriptsize$f_2^3$}};

\draw[thick, ->] (yellow_right) -- (all_back) node[pos=0.15,above=0mm] {{\scriptsize$(f_1^4)^{-1}$}}; 
\draw[thick, ->] (yellow_first) -- (all_back) node[pos=0.15,above=0mm] {{\scriptsize$(f_2^4)^{-1}$}};

\draw[thick, ->] (sky.east) -- (sky_right.west) node[midway,above] {{\scriptsize$I$}};
\draw[thick, ->] (sky_right.east) -- (all_back) node[pos=0.1,above=0mm] {{\scriptsize$I$}};

\end{tikzpicture}
\caption[Generic rectification pipeline]{\textbf{Generic rectification pipeline.} The input image is split into segments, which are processed independently. A set of rectifying transformations is found and applied to each segment. Local features are detected in the transformed segments and their locations are then mapped back by the inverse transformation.} 
\label{fig:general_pipeline}
\end{figure*}
The general idea of image rectification is to undo or reduce the effect of perspective distortion caused by the varying relative position of the camera and the given point in the scene. It is done by applying transformations that produce some kind of normalized view. This can increase the robustness of both feature detector (most importantly in terms of is repeatability and localization precision) and feature descriptor (its matchability). As an example let us assume a surface in the scene with a normal $n$ in point $p$ in the camera coordinates. To provide a description normalized with respect to varying relative position of the camera, the image or its respective part is transformed by a function $f$ so as to simulate a view of the camera looking at $p$ from a given canonical view,. One prominent example of such canonical view is the fronto-parallel view with respect to $n$. Let us assume that after the rectification a keypoint is detected at $f(p)$ and its description $D_{f(p)}$ is computed. To get the rectified keypoint, its description $D_{f(p)}$ is kept, but its position is transformed back by $f^{-1}$, which gives the original position $p=f^{-1}(f(p))$ in the unrectified image. The final rectified keypoint is then given by $(D_{f(p)}, p)$. 

Such formulation of image matching with rectification leaves room for different design choices in each step. Let us compare possible options.

\subsection{The Rectifying Transformation.}
We discuss two main options, which do not produce gaps in synthesized images -- $f$ modeled as a general homography and as an 

$f$ modeled as a general homography accounts for all the transformations mapping $p$ and its small neighborhood (modeled to be locally part of a plane) to a fronto-parallel view. Furthermore, if we assume that $n$ is shared within a larger contiguous region across the depicted surface, this means this whole region is part of one plane. This kind of region can then be rectified all at once. If such a (dominant) plane is identified, the process can be further optimized as only the respective contiguous part of the image need to be rectified at a time. This is advantageous especially for homography rectification as the homography can largely increase the size of the transformed image, increasing the memory footprint and computation complexity of the process. Clipping the region of interest can at least partly solve this problem~\cite{toft}. Additionally, rectifying large portion of image - whether contiguous or not - all at once brings an obvious speed improvement.  

In comparison, affine transformations are only linear approximations of general homographies and hence cannot perform the simulation exactly. However, there are multiple advantages to using affine maps as rectifying transformations to general homographies:
\begin{enumerate}
  \item The resulting image size typically does not grow as big as when using general homographies. This feature can relax the need for search for contiguous regions to be rectified at once, as either the whole image or a general pixel mask can be rectified.   
  \item The qualitative effects of affine rectifications (distance in the space of tilts) as well as the optimization techniques (covering the space of tilts) are very well established~\cite{CoveringSpaceOfTilts}.
  \item Affine transformation allows for directional Gaussian blur which provides anisotropic anti-aliasing and better simulation of the view under different tilt of the camera~\cite{ASIFT}.
\end{enumerate}

\begin{figure*}[htb]
\centering
\begin{tikzpicture}[
every node/.style={inner sep=0,outer sep=2},
node distance = 2mm and 21mm, auto,
rectangularnode/.style={rectangle, draw=black!0, very thin, minimum width=5mm},]

\node[rectangularnode, label={}] (input) {\includegraphics[height=20mm]{imgs/input_image.jpg}};

\node[rectangularnode, label={}] (segments) [right=14mm of input] {\includegraphics[height=20mm]{imgs/frame_out.png}};

\node[rectangularnode, label={[align=center]below:{\small normals clustering}}] (sphere) [below=0mm of segments] {\includegraphics[height=20mm]{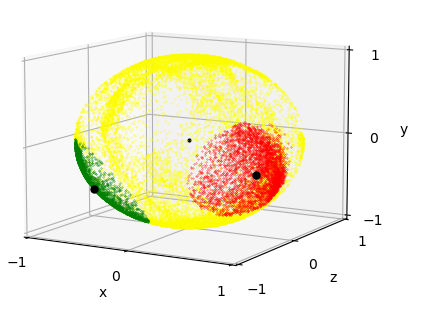}};

\node[rectangularnode, label={}] (red) [right=of segments] {\includegraphics[width=.03\textwidth]{imgs/frame_out_2.png}};
\node[rectangularnode] (green) [above=of red] {\includegraphics[width=.03\textwidth]{imgs/frame_out_3.png}};
\node[rectangularnode] (yellow) [below=of red,distance=2mm and 2mm] {\includegraphics[width=.03\textwidth]{imgs/frame_out_1.png}};
\node[rectangularnode] (sky) [below=of yellow,distance=2mm and 2mm] {\includegraphics[width=.03\textwidth]{imgs/frame_out_sky.png}};

\node[rectangularnode] (green_right) [right=of green,distance=2mm and 2mm] {\includegraphics[width=.03\textwidth]{imgs/frame_out_3_0.png}};

\node[rectangularnode] (red_right) [right=of red,distance=2mm and 2mm] {\includegraphics[width=.03\textwidth]{imgs/frame_out_2_0.png}};

\node[rectangularnode] (yellow_right) [right=of yellow,distance=2mm and 2mm] {\includegraphics[width=.03\textwidth]{imgs/frame_out_y_0.png}};

\node[rectangularnode] (sky_right) [right=of sky,distance=2mm and 2mm] {\includegraphics[width=.03\textwidth]{imgs/frame_out_sky.png}};
\node[rectangularnode] (all_back) [right=of red_right,distance=2mm and 2mm] {\includegraphics[height=20mm]{imgs/frame_out.png}};

\draw[thick, ->] (input) -- (segments) ; 

\draw[thick, ->] (segments) -- (red.west) node[midway,above] {};
\draw[thick, ->] (segments) -- (green.west) node[midway,above] {};
\draw[thick, ->] (segments) -- (yellow.west) node[midway,above] {};
\draw[thick, ->] (segments) -- (sky.west) node[midway,above] {};

\draw[thick, ->] (green.east) -- (green_right.west) node[midway,above] {$H_1$};
\draw[thick, ->] (green_right.east) -- (all_back) node[anchor=left,midway,above=3mm] {$H_1^{-1}$};

\draw[thick, ->] (red.east) -- (red_right.west) node[midway,above] {$H_2$};
\draw[thick, ->] (red_right.east) -- (all_back) node[midway,above] {$H_2^{-1}$};

\draw[thick, ->] (yellow.east) -- (yellow_right.west) node[midway,above] {$H_3$};
\draw[thick, ->] (yellow_right.east) -- (all_back) node[midway,above=1mm] {$H_3^{-1}$};
\draw[thick, ->] (sky.east) -- (sky_right.west) node[midway,above] {$I$};
\draw[thick, ->] (sky_right.east) -- (all_back) node[anchor=left,midway,above=-2mm] {$I$\:\:\:\:\:\:};
\end{tikzpicture}
\caption[Rectification of detected planes via depth maps.]{\textbf{Rectification of detected planes via depth maps.} The image is segmented according to the clusters of normals from the scene depth map. Each segment is rectified by the homography, which aligns the cluster mean to the camera axis and this transforms the segment to the fronto-parallel view.} 
\label{fig:diagram_toft}
\end{figure*}
\subsection{Establishing the Region to Be Rectified.}
Finding (disjoint) parts of the image, so that each of them can be rectified at once, brings a significant speedup as opposed to having to rectify each and every local feature.

Enforcing these parts to be contiguous may correspond to finding (dominant) planes in the scene. This may be beneficial because the structure of the scene is recognized better. Also, transformed contiguous regions will typically be much smaller than the whole transformed image, which may be vital for keeping the method complexity reasonable, e.g. when general homographies are used as the rectification transformations. 

Older methods do not estimate such disjoint parts of the image and rectify the whole image~\cite{CnnAssistedCoverings,ASIFT,MODS} instead. Such methods typically rectify the image by multiple rectifying transformations, which can produce multiple descriptions (each corresponding to one transformation) for a single point.

\subsection{Generic Rectification Pipeline and its Variants}
The high level steps of the rectification pipeline framework are as follows:
\begin{enumerate}
    \item Segment the input image, typically to disjoint parts, potentially to contiguous parts.
    \item Find the set of rectifying transformations for each of the image parts independently. Apply these transformations to the corresponding image parts.     
    \item For each given image part and applied transformation, compute the set of keypoints and their descriptions.
    \item Backproject the keypoints locations by the inverse of their respective rectification transformation.
    \item Output the whole resulting set of features as the union of all the keypoints locations and descriptions from the previous step. 
\end{enumerate}

These steps are depicted in Figure~\ref{fig:general_pipeline}. The framework is general and allows for various design choices in all steps. Some steps are optional and are skipped in some methods, e.g. the segmentation. It is common that some parts of the image are excluded from the rectification altogether and unrectified features are used within these image parts. This is equivalent to an identity being one of the rectifying transformation. 

\begin{figure*}[htb]
\centering
\begin{tikzpicture}[
every node/.style={inner sep=0,outer sep=0},
node distance = 2mm and 32mm, auto,
grode/.style={rectangle, draw=black!0, very thin, minimum width=5mm},]

\node[grode, label={}] (input)
{\includegraphics[height=20mm]{imgs/input_image.jpg}};

\node[grode, label={[align=center]below:{\small covering sparse affine shapes} \\ {\small in the space of tilts}}] [below=of input] (covering)
{\includegraphics[height=20mm]{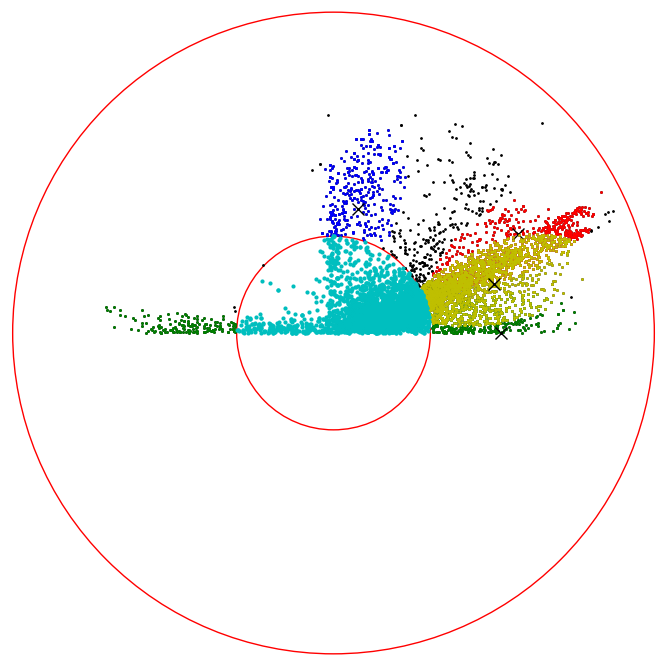}};

\node[grode, label={}] (red) [right=of input] {\includegraphics[width=.06\textwidth]{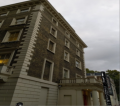}};
\node[grode] (green) [above=of red] {\includegraphics[width=.06\textwidth]{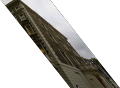}};
\node[grode] (blue) [below=of red,distance=2mm and 2mm] {\includegraphics[width=.06\textwidth]{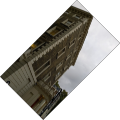}};
\node[grode] (magenta) [below=of blue,distance=2mm and 2mm] {\includegraphics[width=.06\textwidth]{imgs/input_image.jpg}};

\node[grode] (all_back) [right=of red,distance=2mm and 2mm] {\includegraphics[height=20mm]{imgs/input_image.jpg}};

\draw[thick, ->] (input) -- (red.west) node[midway,above=1mm] {{\footnotesize $A_2$}};
\draw[thick, ->] (input) -- (green.west) node[midway,above=1mm] {{\footnotesize $A_1$}};
\draw[thick, ->] (input) -- (blue.west) node[midway,above=1mm] {{\footnotesize $A_3$}};
\draw[thick, ->] (input) -- (magenta.west) node[midway,above=2mm] {{\footnotesize $I$}};

\draw[thick, ->] (red.east) -- (all_back) node[midway,above=1mm] {{\footnotesize $A_2^{-1}$}};
\draw[thick, ->] (green.east) -- (all_back) node[midway,above=1mm] {{\footnotesize $A_1^{-1}$}};
\draw[thick, ->] (blue.east) -- (all_back) node[midway,above=1mm] {{\footnotesize $A_3^{-1}$}};
\draw[thick, ->] (magenta.east) -- (all_back) node[midway,above=0mm] {{\footnotesize $I$\:\:\:\:\:\:}};

\end{tikzpicture}
\caption[Rectification via covering the AffNet shapes]{\textbf{Rectification via covering the AffNet shapes in the space of tilts.} No segmentation is done here. A set of rectifying affine maps is found, so that it covers at least some thresholded ratio of the AffNet shapes in the space of tilts. These shapes are sparse, meaning they are computed only in the keypoint locations. The predecessor of this method, ASIFT, covered the whole space of tilts (up to some maximal tilt value, which corresponds to a certain maximal distance from the origin in the space of tilts) with a fixed set of (typically 61) rectifying affine maps.}
\label{fig:diagram_rodriguez}
\end{figure*}

\paragraph{Repeat-based Rectification.}
Variants of this method are presented in series of works~\cite{RepPatRect2018a,RepPat2018,pritts2019minimal,pritts2020}. A main of assumption of such methods, that there is some repeated pattern, presented in the image, e.g. many similar looking windows. If one can successfully detect and match those repeats, then from the difference in their geometry, perspective and even radial distortions could be estimated and rectified. While such methods are powerful, they often are brittle and rarely used~\cite{toft}. 
\paragraph{Simple Depth-map-based Rectification.}
Idea to use depth information to get both segmentation and rectified transformation is first proposed in~\cite{toft}.%
The image depth map can be obtained via depth estimating models like MonoDepth~\cite{monodepth2} or MegaDepth~\cite{MDLi18}. The segmentation is done via clustering of the detected normals of the scene on a unit sphere. 
The clusters of normals are found by voting among a set of buckets corresponding to N points (typically N=500) distributed roughly equidistantly on the unit sphere~\cite{SpiralingAlgorithm}. A greedy algorithm finds iteratively buckets with the most votes as long as they have some minimal number of votes. The clusters centers can be refined either by mean-shift algorithm or by simply adjusting to the mean of the cluster. In the original method~\cite{toft} the orthogonality of the cluster centers, which are estimates of plane normals, is enforced. This somehow limits the generality of the method. In the next step, image segments are found as connected components of pixels, whose normals belong to the same cluster and which cover at least some minimal image area. 

The rectification is done via homographies, one per segment. %
This method has multiple hyperparameters related to the normals clustering, plane orthogonality enforcement, handling of antipodal points in the normal space, etc.%
A re-implementation of this method is used for the experiments here~\footnote{Authors' implementation with custom monocular depth model is not available, thus we did our best in re-implementing it}. The benefit of this method is that it rectifies image regions directly based on the geometry of the scene. As it rectifies only the detected planes, the rectified parts of images can be cropped, which is a performance improvement over other homography rectifying methods. Large planes are often found in many real world scenes like street views. In scenes where no large planes are present this method does not perform particularly well, even though it gracefully degrades to keeping the features unrectified. Another advantage is that this method utilizes monocular depth estimation models, which are receiving significant research attention now~\cite{AdaBins2021}. Potentially, it may automatically improve with the progress in monocular depth estimation. Finally, the rectification maps find by this method are geometrically meaningful and potentially can be re-used for other tasks. The depth-map-based method is depicted in Figure~\ref{fig:diagram_toft}.

\begin{figure}[htb]
\centering
{{\includegraphics[width=0.8\linewidth]{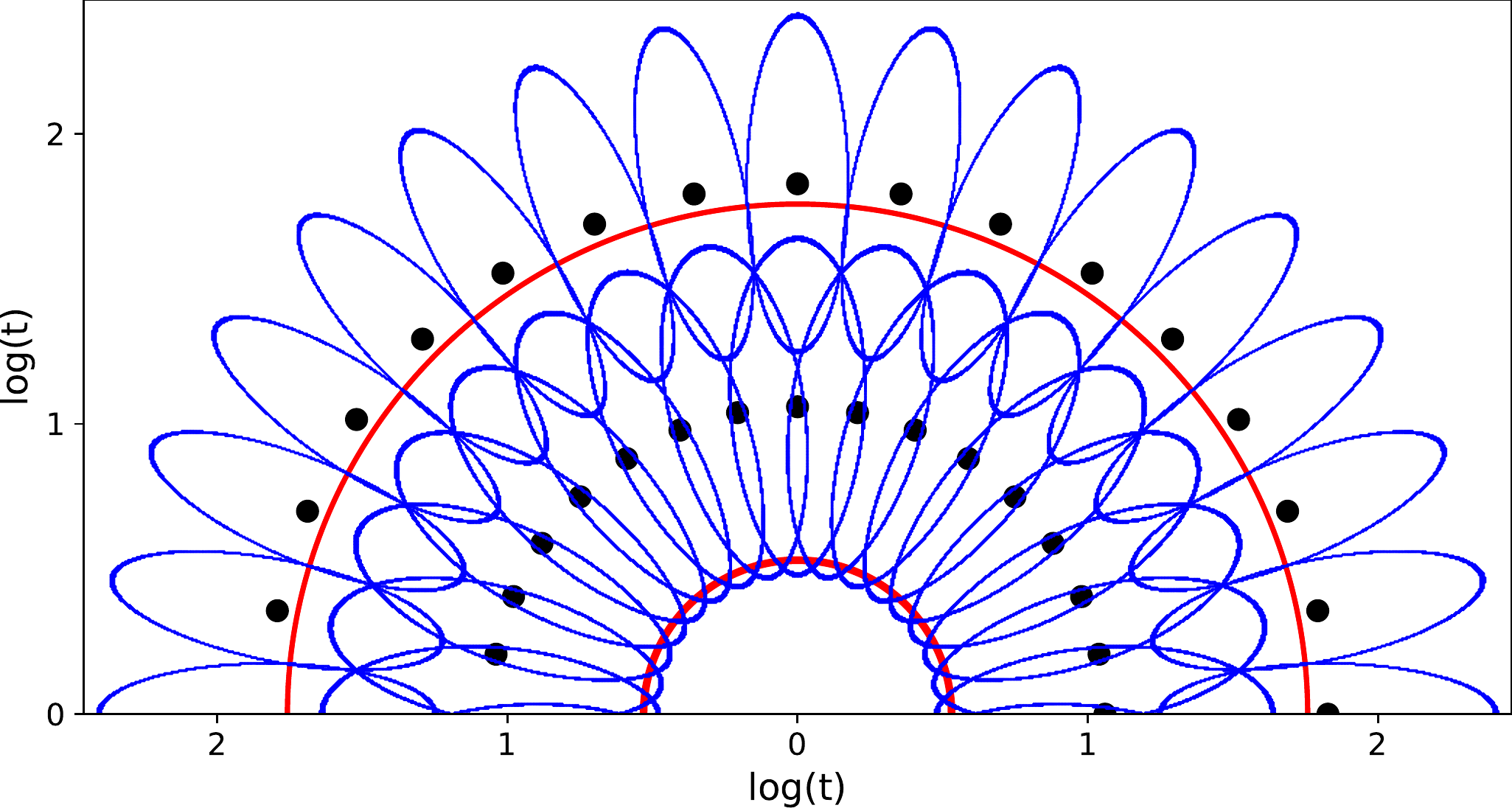}}}
\caption{\textbf{Space of Tilts} is a quotient space of affine maps. Logarithm of affine tilt and angle $\phi$ from the affine map decomposition~\cite{CnnAssistedCoverings} are radius and angle in the polar coordinate system respectively. The metric in this space induce circular r-balls, if centered in origin, which are more and more prolonged if centered further from the origin (red half-circle and blue ellipse-like shapes with black centers). This (semi-)metric treats points $(t_0, \phi)$ and $(t_0, \phi + \pi)$ as equivalent (distance=0). The covering covers the space of tilts up to some maximal tilt (larger red half-circle). }
\label{fig:sot}
\end{figure}

\paragraph{Simple AffNetShapes Rectification.}
This method uses local feature affine shape estimation, like AffNet, to estimate multiple rectifying transformation for the whole image~\cite{CnnAssistedCoverings}. It is based on findings from the work on ASIFT~\cite{ASIFT} and the space of tilts~\cite{CoveringSpaceOfTilts}. In ASIFT, a large fixed set of 61 rectification transformations is applied to the image. These transformations correspond to centers of minimal covering r-balls in the space of tilts, which is a quotient space of affine maps (see Figure~\ref{fig:sot}). The r-balls common radius r is set so that the covered affine maps are close enough with respect to (SIFT) keypoints matchability and repeatability.
In the follow-up work, which we will call simple AffNetShapes rectification~\cite{CnnAssistedCoverings}, a smaller dynamic set of rectifying transformations is found with the help of affine shapes estimates obtained from AffNet CNN~\cite{AffNet} in the detected keypoint locations. A greedy algoritm is used to find this set, so that the corresponding r-balls in the space of tilts cover at least some minimal portion of the found AffNet shapes (typically a value of $0.95$ is used).
This lowers a computation complexity of the rectification. The set of rectifying affine maps are all applied to the whole image, as in ASIFT. Local features from the original image are kept as well.%

In contrast to the rectification via normals, this method may not reflect the scene geometry, as AffNet shapes are originally used to normalize patches for HardNet or SIFT descriptor, irrespective of whether it actually transforms the patch to a fronto-parallel or some other fixed canonical view.
The benefit of using this method compared to just applying AffNet to rectify local patches for description is that when the transformation is used prior to keypoint detection, it allows to detect new keypoints, which could not be detected in the original image.
This method is depicted in Figure~\ref{fig:diagram_rodriguez}.

\begin{figure*}[htb]
\centering
\begin{tikzpicture}[
every node/.style={inner sep=2,outer sep=2},
node distance = 2mm and 24mm, auto,
grode/.style={rectangle, draw=black!0, very thin, minimum width=5mm},
]

\node[grode, label={}] (input) {\includegraphics[height=20mm]{imgs/input_image.jpg}};

\node[grode, label={}] (input_sgm) [right=10mm of input] {\includegraphics[height=20mm]{imgs/frame_out.png}};

\node[grode, label={[align=center]left:{\small normals clustering}}] (sphere) [below=-1mm of input_sgm] {\includegraphics[height=20mm]{imgs/sphere.png}};

\node[grode, label={}] (red) [right=of input_sgm] {\includegraphics[width=.03\textwidth]{imgs/frame_out_2.png}};
\node[grode] (green) [above=-2mm of red] {\includegraphics[width=.03\textwidth]{imgs/frame_out_3.png}};
\node[grode] (yellow) [below=-2mm of red,distance=2mm and 2mm] {\includegraphics[width=.03\textwidth]{imgs/frame_out_1.png}};
\node[grode] (sky) [below=of yellow,distance=2mm and 2mm] {\includegraphics[width=.03\textwidth]{imgs/frame_out_sky.png}};

\node[grode, label={[align=center]left:{\small covering sparse affine shapes} \\ {\small in the space of tilts}}] [above=-1mm and -0mm of green] (covering)
{\includegraphics[height=20mm]{imgs/covering_basic.png}};

\node[grode] (green_right) [right=of green] {\includegraphics[width=.03\textwidth]{imgs/frame_out_3_1.png}};
\node[grode] (green_first) [above=of green_right,distance=2mm and 2mm] {\includegraphics[width=.03\textwidth]{imgs/frame_out_3_0.png}};
\node[grode] (green_second) [above=of green_first,distance=2mm and 2mm] {\includegraphics[width=.03\textwidth]{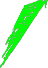}};
\node[grode] (green_third) [below=of green_right,distance=2mm and 2mm] {\includegraphics[width=.03\textwidth]{imgs/frame_out_3_3.png}};

\node[grode] (red_right) [below=-1mm of green_third,distance=2mm and 2mm] {\includegraphics[width=.03\textwidth]{imgs/frame_out_2_0.png}};
\node[grode] (red_first) [below=-1mm of red_right,distance=2mm and 2mm] {\includegraphics[width=.03\textwidth]{imgs/frame_out_2_1.png}};

\node[grode] (yellow_right) [below=-1mm of red_first,distance=2mm and 2mm] {\includegraphics[width=.03\textwidth]{imgs/frame_out_y_0.png}};

\node[grode] (sky_right) [right=of sky,distance=2mm and 2mm] {\includegraphics[width=.03\textwidth]{imgs/frame_out_sky.png}};

\node[grode] (all_back) [right=13cm of input] {\includegraphics[height=20mm]{imgs/frame_out.png}};

\draw[thick, ->] (input) -- (input_sgm) node[midway,above] {};

\draw[thick, ->] (input_sgm) -- (red.west) node[midway,above] {};
\draw[thick, ->] (input_sgm) -- (green.west) node[midway,above] {};
\draw[thick, ->] (input_sgm) -- (yellow.west) node[midway,above] {};
\draw[thick, ->] (input_sgm) -- (sky.west) node[midway,above] {};

\draw[thick, ->] (green) -- (green_right.west) node[midway,above] {{\footnotesize$A_3^1$}};
\draw[thick, ->] (green) -- (green_first.west) node[midway,above] {{\footnotesize$A_2^1$}};
\draw[thick, ->] (green) -- (green_second.west) node[midway,above] {{\footnotesize$A_1^1$}};
\draw[thick, ->] (green) -- (green_third.west) node[midway,above] {{\footnotesize$A_4^1$}};

\draw[thick, ->] (green_right) -- (all_back) node[pos=0.1,above=1mm] {{\footnotesize$(A_3^1)^{-1}$}};
\draw[thick, ->] (green_first) -- (all_back) node[pos=0.1,above=2mm] {{\footnotesize$(A_2^1)^{-1}$}};
\draw[thick, ->] (green_second) -- (all_back) node[pos=0.1,above=3mm] {{\footnotesize$(A_1^1)^{-1}$}};
\draw[thick, ->] (green_third) -- (all_back) node[pos=0.1,above=0mm] {{\footnotesize$(A_4^1)^{-1}$}};

\draw[thick, ->] (red) -- (red_right.west) node[midway,above=-1mm] {{\footnotesize$A_1^2$}};
\draw[thick, ->] (red) -- (red_first.west) node[midway,above=-1mm] {{\footnotesize$A_2^2$}};

\draw[thick, ->] (red_right) -- (all_back) node[pos=0.1,above=-1mm] {{\footnotesize$(A_1^2)^{-1}$}};
\draw[thick, ->] (red_first) -- (all_back) node[pos=0.1,above=-1mm] {{\footnotesize$(A_2^2)^{-1}$}};

\draw[thick, ->] (yellow) -- (yellow_right.west) node[midway,above=-1mm] {{\footnotesize$A_1^3$}};

\draw[thick, ->] (yellow_right) -- (all_back) node[pos=0.1,above=-1mm] {{\footnotesize$(A_1^3)^{-1}$}};

\draw[thick, ->] (sky.east) -- (sky_right.west) node[midway,above] {{\footnotesize$I$}};
\draw[thick, ->] (sky_right.east) -- (all_back) node[pos=0.1,above=-1mm] {{\footnotesize$I$}};

\end{tikzpicture}
\caption[DepthAffNet rectification]{\textbf{DepthAffNet rectification.} The image is segmented according to clusters of the normals from a depth map. Then each segment is processed independently. Rectification via covering the AffNet shapes is performed on each of them.} 
\label{fig:diagram_depth_affnet}
\end{figure*}

\paragraph{DepthAffnet rectification.}

This is the first of the two methods we propose.
It combines the two previous methods - simple depth-map-based and simple AffNetShapes rectification.
The segmentation is done via clustering of the normals obtained from a depth estimating CNN~\cite{monodepth2,MDLi18}, just as in simple depth-map-based rectification. The detected contiguous parts of the image are each fed as the input to the AffNetShapes rectification~\cite{CnnAssistedCoverings}. Different segments show different distributions of shapes in the space of tilts - not only due to the different dominating normal direction, but possibly also due to structural alignment (bricks in walls).%
Additionally, it requires smaller portions of the image to be rectified.
This method is depicted in Figure~\ref{fig:diagram_depth_affnet}.

\begin{figure*}[htb]
\centering

\begin{tikzpicture}[
every node/.style={inner sep=2,outer sep=2},
node distance = 2mm and 25mm, auto,
grode/.style={rectangle, draw=black!0, very thin, minimum width=5mm},
rectangularnode/.style={rectangle, draw=black!60, very thick, minimum height=2mm},]

\node[grode, label={}] (input) {\includegraphics[height=20mm]{imgs/input_image.jpg}};

\node[grode, label={}] (input_sgm) [right=8mm of input] {\includegraphics[height=20mm]{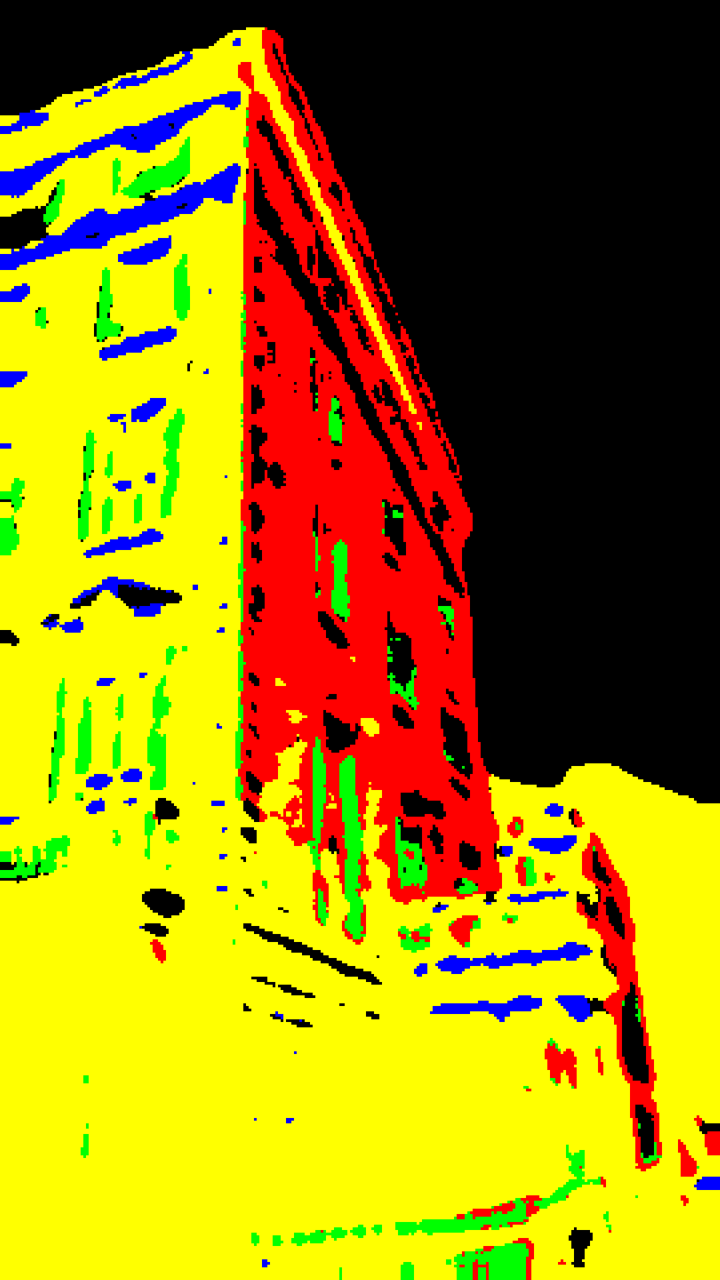}};

\node[grode, label={[align=center]below:{\small dense affine shapes components}}] (sphere) [below left=5mm and -15mm of input_sgm] {\includegraphics[height=15mm]{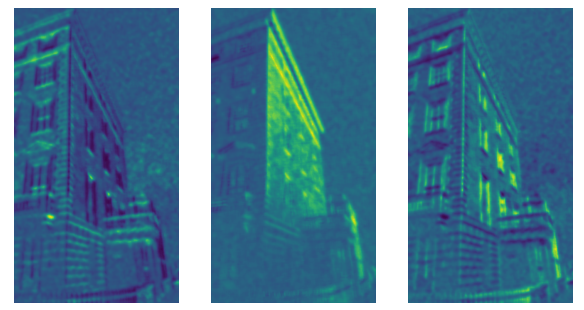}};

\node[grode, label={[align=center]above:{\small covering dense affine shapes} \\ {\small in the space of tilts}}] [above left=1mm and -15mm of input_sgm] (covering)
{\includegraphics[height=20mm]{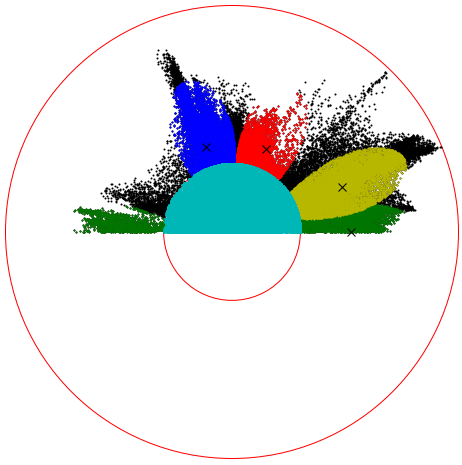}};

\node[grode, label={}] (red) [right=of input_sgm] {\includegraphics[height=14mm]{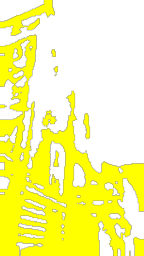}};
\node[grode] (green) [above=-2mm of red] {\includegraphics[height=14mm]{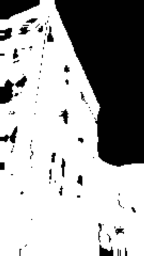}};
\node[grode] (yellow) [above=of green,distance=2mm and 2mm] {\includegraphics[height=14mm]{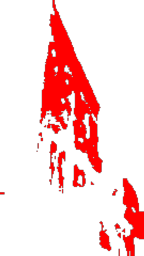}};
\node[grode] (magenta) [below=of red,distance=2mm and 2mm] {\includegraphics[height=14mm]{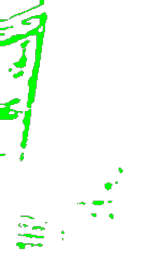}};
\node[grode] (sky) [below=of magenta,distance=2mm and 2mm] {\includegraphics[height=14mm]{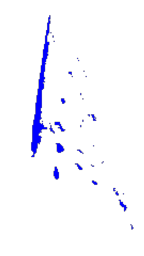}};

\node[grode] (red_right) [right=of red] {\includegraphics[height=14mm]{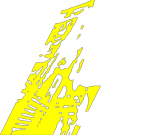}};

\node[grode] (green_right) [right=of green] {\includegraphics[height=14mm]{imgs/segments_dense_affnet_3.png}};

\node[grode] (yellow_right) [right=of yellow] {\includegraphics[height=14mm]{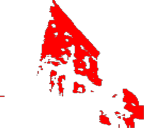}};

\node[grode] (magenta_right) [right=of magenta] {\includegraphics[height=14mm]{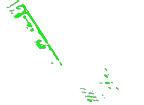}};

\node[grode] (sky_right) [right=of sky] {\includegraphics[height=14mm]{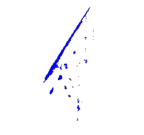}};

\node[grode] (all_back) [right=of red_right] {\includegraphics[height=20mm]{imgs/frame_0000001350_2.png}};

\draw[thick, ->] (input) -- (input_sgm) node[midway,above] {};

\draw[thick, ->] (input_sgm) -- (red.west) node[midway,above] {};
\draw[thick, ->] (input_sgm) -- (green.west) node[midway,above] {};
\draw[thick, ->] (input_sgm) -- (yellow.west) node[midway,above] {};
\draw[thick, ->] (input_sgm) -- (magenta.west) node[midway,above] {};
\draw[thick, ->] (input_sgm) -- (sky.west) node[midway,above] {};

\draw[thick, ->] (green) -- (green_right.west) node[midway,above=1mm] {$I$};
\draw[thick, ->] (green_right) -- (all_back) node[midway,above=1mm] {$I$};

\draw[thick, ->] (red) -- (red_right.west) node[midway,above] {$A_2$};
\draw[thick, ->] (red_right) -- (all_back) node[pos=0.42,above] {$A_2^{-1}$};

\draw[thick, ->] (yellow) -- (yellow_right.west) node[midway,above] {$A_1$};
\draw[thick, ->] (yellow_right) -- (all_back) node[pos=0.4,above=2mm] {$A_1^{-1}$}; 

\draw[thick, ->] (magenta) -- (magenta_right.west) node[midway,above] {$A_3$};
\draw[thick, ->] (magenta_right) -- (all_back) node[pos=0.27,above=0mm] {$A_3^{-1}$}; 

\draw[thick, ->] (sky.east) -- (sky_right.west) node[midway,above] {$A_4$};
\draw[thick, ->] (sky_right.east) -- (all_back) node[pos=0.37,above=2mm] {$A_4^{-1}$};

\end{tikzpicture}
\caption[DenseAffnet rectification]{\textbf{DenseAffnet rectification.} Dense AffNet is run on the input image, which is split into sets according to the r-balls covering AffNet shapes. These sets are generic disjoint image masks. Each of these are then rectified by a single affine map.} 
\label{fig:diagram_dense_affnet}
\end{figure*}

\paragraph{DenseAffnet rectification.}
This is the second method proposed here. It takes a step further and replaces the depth-map-based segmentation with the AffNet-based one, thus reducing number of components of the method.
Unlike previous methods, where AffNet was used to estimate affine shape of the local feature at keypoint location, AffNet is run densely here. We modify (without retraining) AffNet from patch-based to be fully-convolutional model, which predicts an affine shape for every 4x4 pixel block in the image. A similar modification was done in~\cite{D2D2020} to convert HardNet and SOSNet patch descriptors to dense ones. 
When covering the space of tilts with the r-balls corresponding to affine transformations, the covered affine shapes define, through their locations in the image, disjunctive image regions, each of which can be rectified by a single affine transformation. Every point in image will thus correspond to at most one rectified local feature. On top of that features from the original image are kept as well as with the previous affine rectifying methods. We shown that this method is faster than DepthAffnet rectification, as less rectification transformations are performed and the keypoint detection is restricted to the mask of the respective image region. Furthermore, in scenes without many large planes the accuracy of DenseAffnet rectification is even bigger than the state-of-the-art. The method is depicted in Figure~\ref{fig:diagram_dense_affnet}.

\section{Experiments and results}
\begin{figure*}[htb]
\centering
\includegraphics[width=0.31\linewidth]{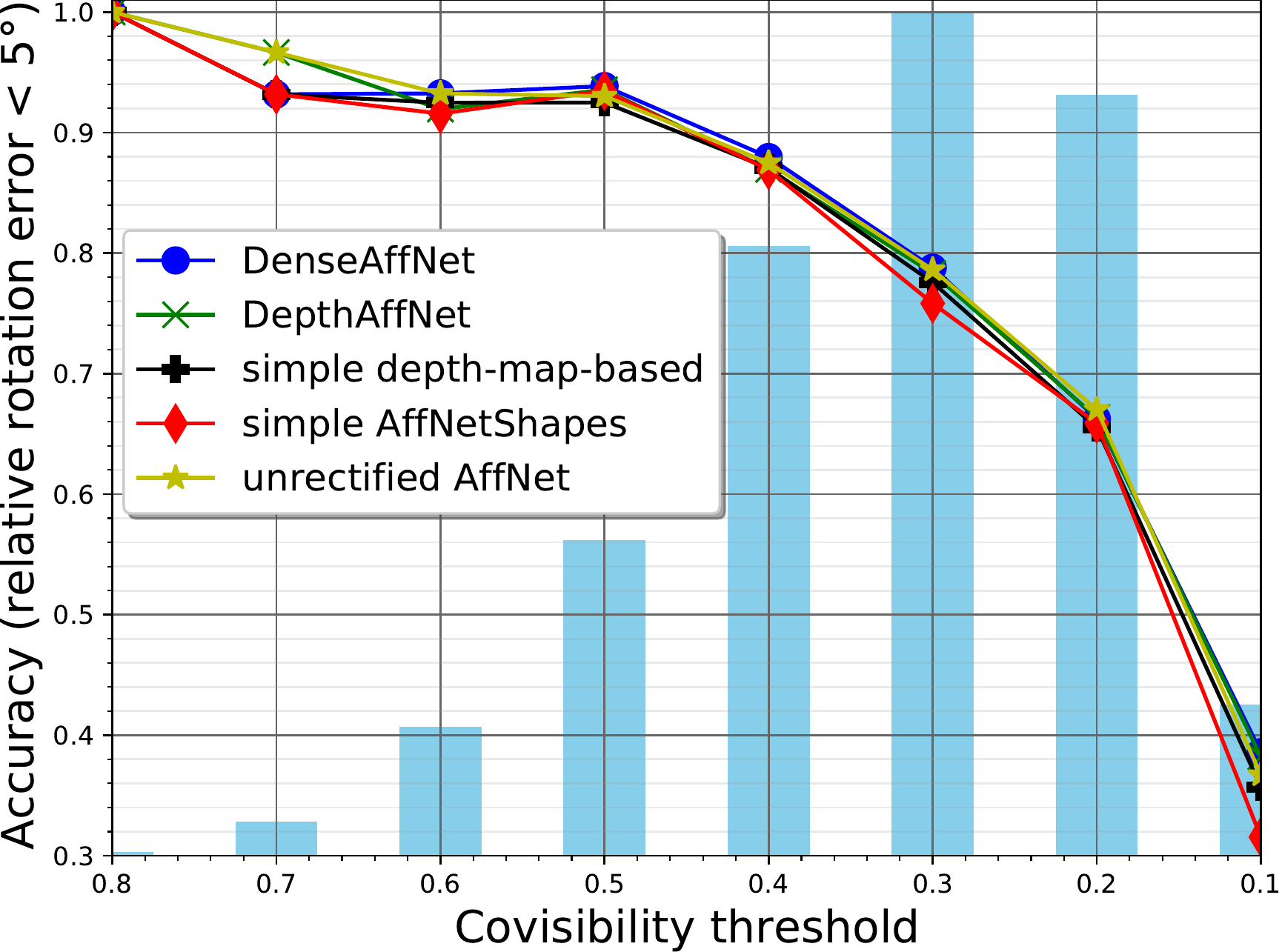}
\includegraphics[width=0.31\linewidth]{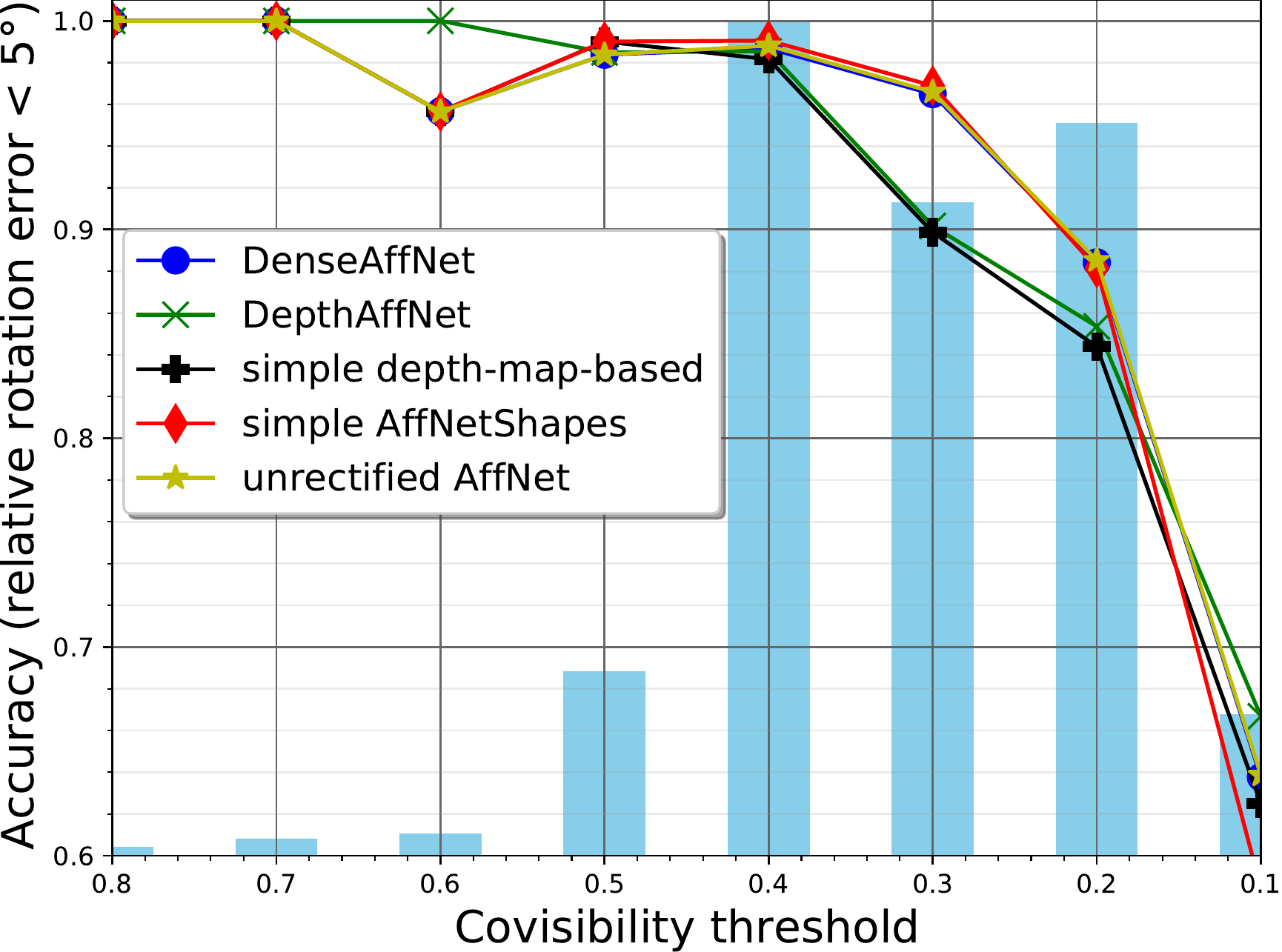}
\includegraphics[width=0.31\linewidth]{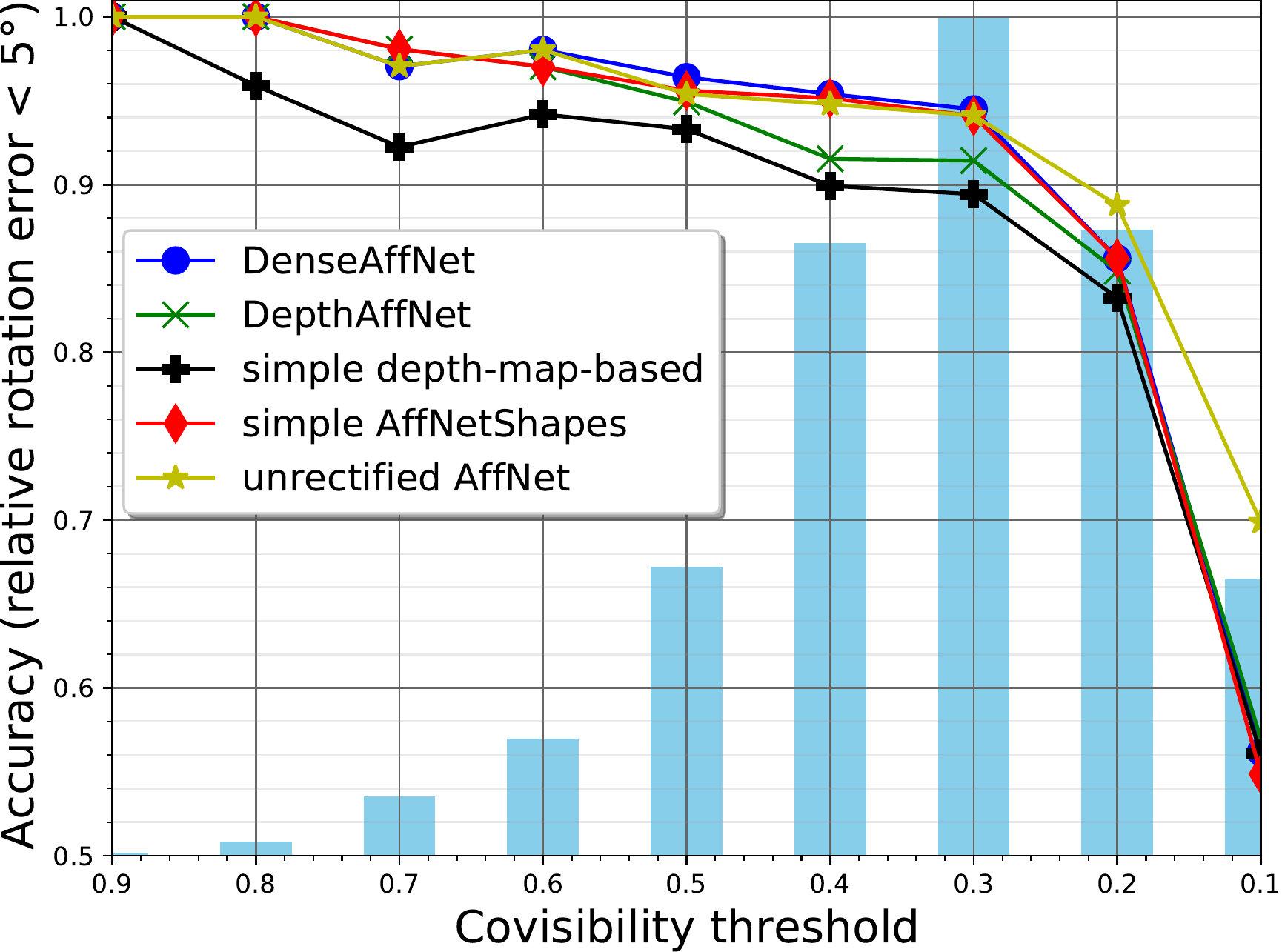}
\caption{\textbf{IMC Phototourism dataset results: average accuracy of camera pose}. From left to right: St. Peter's Square, Sacre Coeur and Reichstag. Vertical bars show the relative sizes of different categories corresponding to different relative covisibility thresholds. Results for DenseAffNet, DepthAffNet, simple depth-map-based~\cite{toft} and simple AffNetShapes~\cite{CnnAssistedCoverings} rectification methods are shown as well as for unrectified HardNet~\cite{HardNet} descriptions normalized by ~\cite{AffNet} using DoG detector.}
\label{fig:accuracy_IMC}
\end{figure*}
\subsection{Datasets}
\paragraph{Strong ViewPoint Changes Dataset.}
This dataset~\cite{toft} captures outdoor scenes from a city, near the building wall. It  consist of 8 scenes, each containing images of one building taken from a wide range of different viewpoints. For each scene, the set of image pairs is split in up to 18 different categories depending on their relative rotations - the pairs in the $k$-th category (called difficulty) have relative rotation in the range
of $[10k, 10(k + 1))$ degrees. Each scene's difficulty consists of up to 200 pairs of images. 
We have used Scene 1 as a "training set", i.e. we have tuned various hyper-parameters using results on Scene 1. The rest of the scenes and datasets are used as test set, i.e. all the experiments were run only once there.
\paragraph{IMC Phototourism dataset.}
IMC Phototourism consists of various outdoor landmarks, reconstructed from many tourist photos, hence the name. We have used the validation set scenes were used: St.Peter's square, Sacre Coeur and Reichstag~\cite{IMC2020}. Each scene is split into categories with different relative covisibility area of their image pairs. A single category contains image pairs with relative covisibility area in the range of $[0.1k, 0.1(k + 1))$ for $k \in \{1,2, \dots 9\}$. Category sizes vary rather greatly from under 10 image pairs to almost 1500 image pairs in a single category.

In both IMC PT and SVP datasets, ground truth comes from the Colmap reconstruction.

\paragraph{The EVD dataset.}
This dataset was introduced in the work on MODS~\cite{MODS} and provides 15 image pairs with extreme viewpoint change and a single dominating plane. 

\subsection{Performance metric}
\begin{figure*}[htb]
\centering
\includegraphics[width=0.4\linewidth]{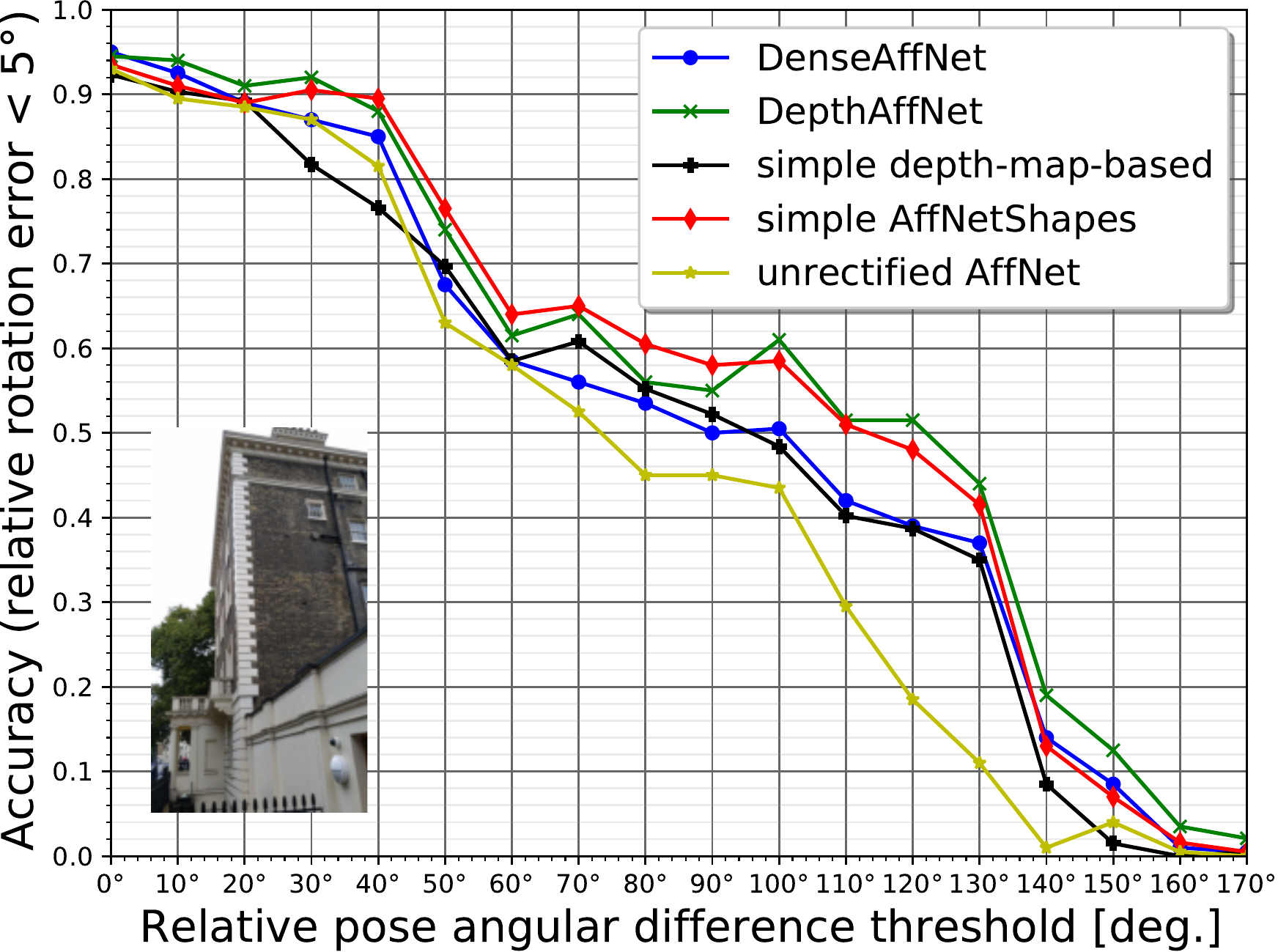}
\includegraphics[width=0.4\linewidth]{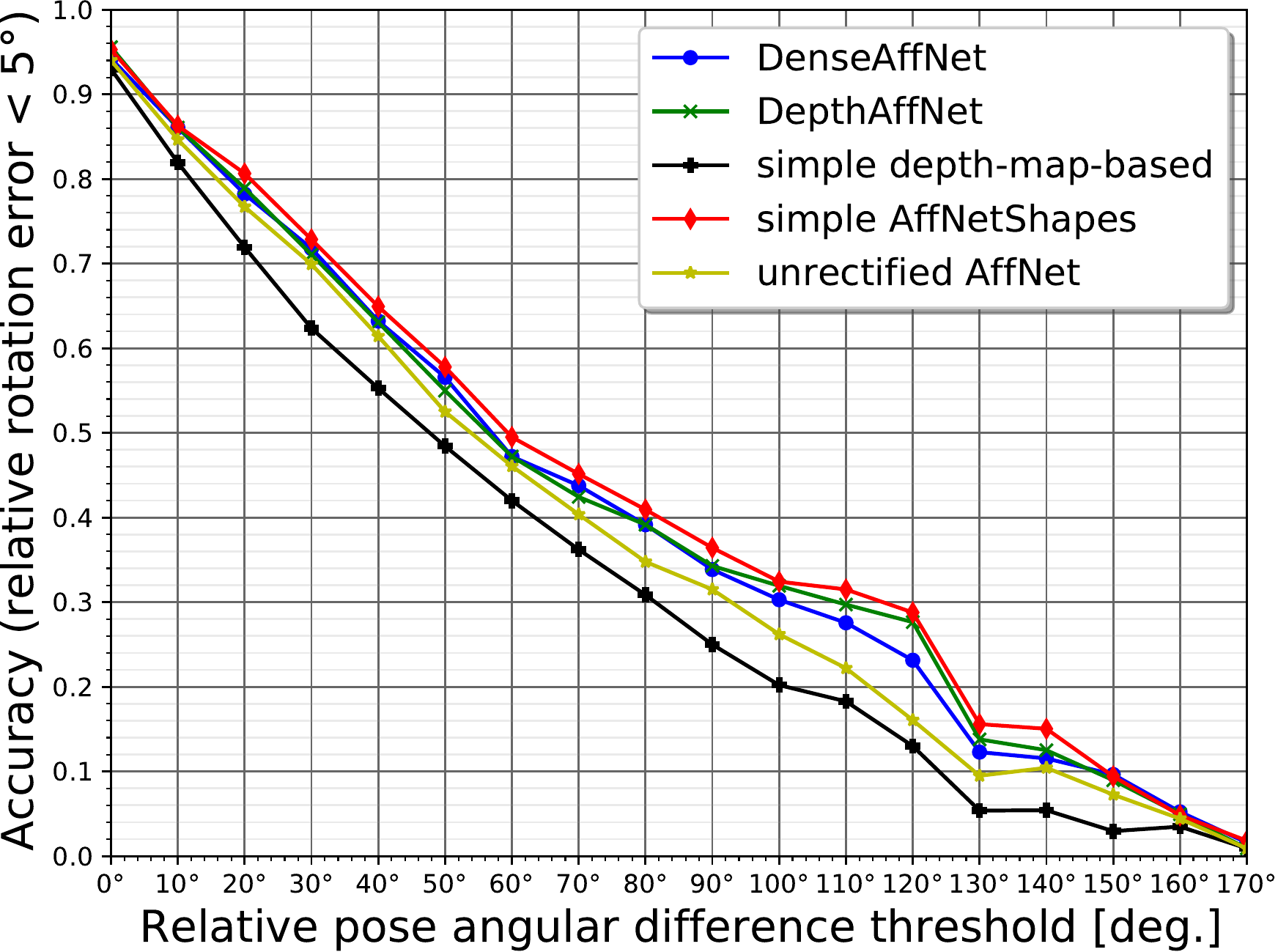}
\caption{\textbf{Accuracy on scene 1 (left) and  scenes 2 to 8 (right)  of the Strong ViewPoint Changes Dataset.} Scene 1 is used as a training set, 2 -- 8 as a test set. The results are shown for DenseAffNet, DepthAffNet, simple depth-map-based~\cite{toft} and simple AffNetShapes~\cite{CnnAssistedCoverings} rectification methods as well for unrectified HardNet~\cite{HardNet} descriptions normalized by ~\cite{AffNet} using DoG detector.}  
\label{fig:accuracy_scene1}
\end{figure*}
For the first two datasets, the performance metric was the accuracy of the estimated relative camera rotation between the image pair using rectified keypoints from the given method. 
The pose estimation was computed via OpenCV RANSAC~\cite{RANSAC1981} implementation~\cite{itseez2015opencv} (cv2.RANSAC flag) using the known ground truth camera intrinsics, estimating the essential matrix. The pose estimate was considered accurate, if the difference between the estimated relative rotation and the ground truth was smaller than 5$^{\circ}$. For each experiment the ratio of accurately estimated image pairs was measured per scene(s) and category. This is exactly the same performance metric that was used in~\cite{toft}. 
Note, that we deviate from the metric used in IMC paper~\cite{IMC2020} to be more consistent with comparison used in depth-map-based paper~\cite{toft}. One more reason to do it, is that IMC metric uses uncalibrated setup and fundamental matrix estimation algorithm is more sensitive to having many in-plane matches, unlike essential matrix estimator which is out of the scope of current paper.

Unlike by the first two datasets, by the EVD dataset a homography is to be estimated for each image pair and the performance metric is the number of pairs with the mean absolute reprojection error of the visible part lower than given threshold values (1, 2, 3, 5, 10 and 20 pixels respectively).

Parameter values for RANSAC were set for both essential matrix and homography estimation as follows: threshold=0.5 pixels, confidence=$1-$$10^{-4}$, iterations=$10^5$.

Unlike~\cite{toft}, which used SIFT~\cite{SIFT} we decide to use DoG-AffNet-HardNet~\cite{AffNet,HardNet} local feature as an unrectified baseline, as implemented in kornia~\cite{kornia}. The reason is that DoG-AffNet-HardNet performed among the top entries in IMC benchmark~\cite{IMC2020} and is much more robust to viewpoint changes than vanilla SIFT.
\subsection{Results}
\begin{figure*}[htb]
\centering
\includegraphics[width=0.84\linewidth]{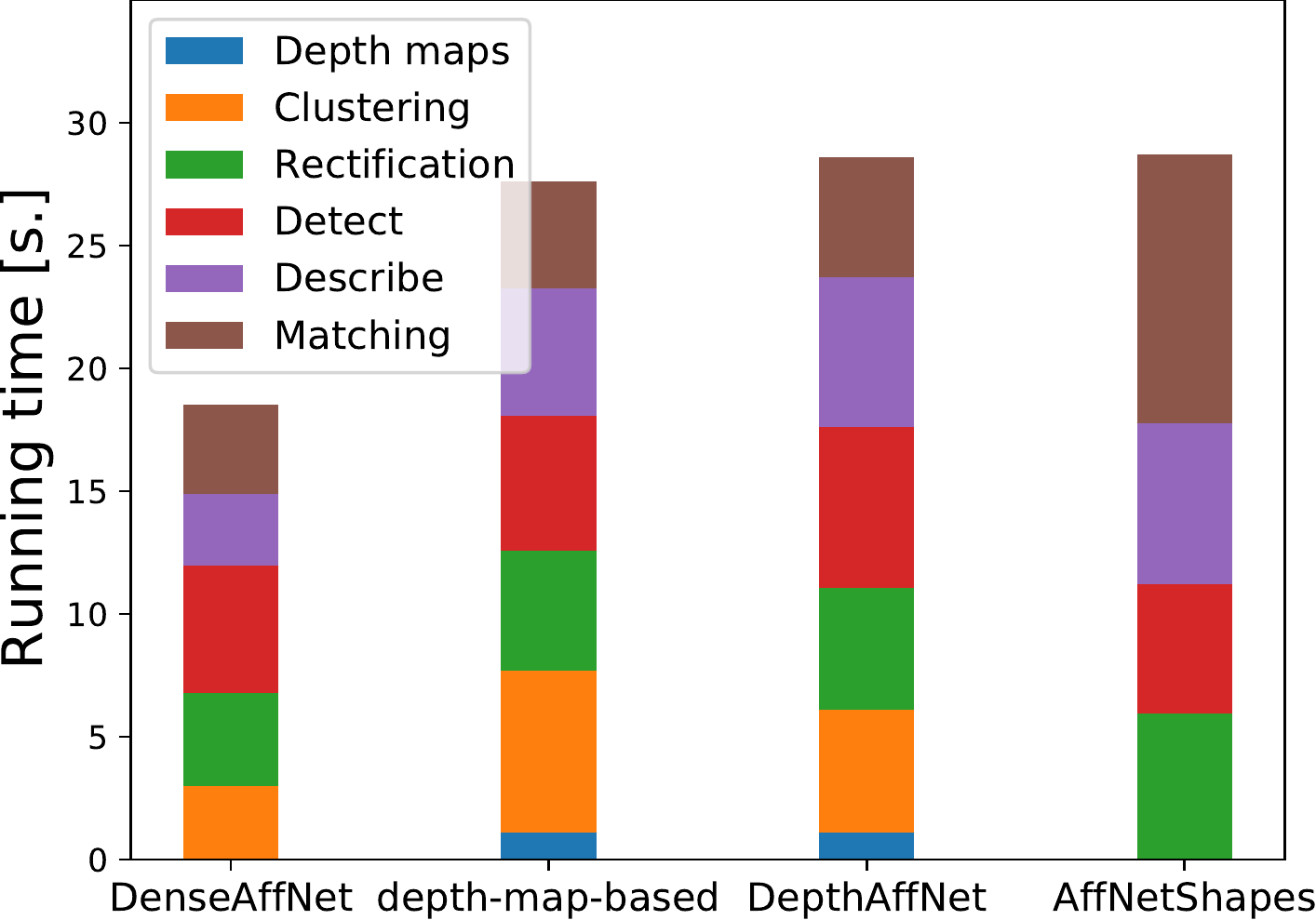}
\caption{Running times of different components of DenseAffNet, DepthAffNet, simple depth-map-based~\cite{toft} and simple AffNetShapes~\cite{CnnAssistedCoverings} rectification methods. Times measured as average running time per image on scene 1 of the Strong ViewPoint Changes.}
\label{fig:running_time}
\end{figure*}
\begin{table}[htb]
\footnotesize
\begin{tabular}{ccccc}
\toprule
&  area [$10^6$px] & kpts [K] & \#comp & rect/comp \\ \midrule
DenseAffNet & 2.98 & 24 & 2.73 & 1.00 \\ 
DepthAffNet & 3.92 & 35& 1.88 & 3.61 \\ AffNetShapes~\cite{CnnAssistedCoverings} & 6.16 & 52 & 1.00 & 5.35 \\ 
depth-map-based~\cite{toft} & 1.72 & 18 & 1.89 & 1.00 \\ 
\bottomrule
\end{tabular}
\caption{\textbf{Complexity statistics for DepthAffNet and DenseAffNet.} Comp. - component, rect - rectification.  Rectified image parts are cropped to produce the smallest rectangular image possible. This is not as efficient for generic image masks in DenseAffNet as for contiguous segments in DepthAffNet. However, the fact that DepthAffNet performs multiple rectifications makes DepthAffNet produce warps with larger total area per image and detect more keypoints. All data are average values measured on scene 1 of the Strong ViewPoint Changes Dataset. The numbers of keypoints are rounded to hundreds.}
\label{table:complexity_stats}
\end{table}
Results on the Strong ViewPoint Changes Dataset are shown in Figure~\ref{fig:accuracy_scene1}. One can see that the gap between unrectified and rectified methods with use of modern local features is much less pronounced than reported in~\cite{toft}. Nevertheless, rectification improves results, especially for the more difficult cases, with AffNet-based methods on the top. Among AffNet-based methods, DenseAffNet is the fastest (see Figure~\ref{fig:running_time} and Table~\ref{table:complexity_stats}).
 IMC Phototourism dataset (Figure~\ref{fig:accuracy_IMC} contains smaller viewpoint changes than SVP dataset, and unrectified local feature together with proposed DenseAffNet shows the best results. Simple depth-based~\cite{toft} and AffNet-based~\cite{CnnAssistedCoverings} perform a bit worse, than unrectified version.
 
On EVD dataset the simple affine rectification~\cite{CnnAssistedCoverings} performed the best, followed closely by DepthAffnet. DenseAffnet had a bit worse performance, yet better than the simple depth-map-based rectification~\cite{toft} (see Table~\ref{table:EVD}). This dataset presents rather a special case with a single dominant plane and extreme (hence the name) viewpoint changes.
Summarizing, DenseAffnet and DepthAffnet show consistent performance across different datasets without degradation in cases of smaller viewpoint change present. They are also on-par or faster then previously proposed methods. 
\begin{table*}[htb]
\begin{center}
{\footnotesize{
\begin{tabular}{  c c  c  c  c  c } 
 \hline
MAE th.[px] & DenseAffNet & DepthAffNet & simple AffNetShapes~\cite{CnnAssistedCoverings} & simple depth-map-based~\cite{toft} & unrectified AffNet \\
\hline
1 &	0 &	0 & 0 & 0 & 0\\
\hline
2 &	1 &	1 &	2 &	0 & 0\\
\hline
3 &	3 &	2 &	5 &	0 & 1\\
\hline
5 &	3 &	4 &	8 &	1 & 2\\
\hline
10 & 3 & 7 & 9 & 2 & 2\\
\hline
20 & 5 & 9 & 10 & 2 & 4\\
\hline
time[s.] & 114.73 &	175.04 & 185.47 & 169.92 & 35.98 \\
\end{tabular}
}}
\end{center}
\caption[Results on EVD~\cite{MODS}]{\textbf{Results on Extreme View Dataset.} Number of pairs (out of 15), with mean absolute reprojection error below the threshold.} 
\label{table:EVD}
\end{table*}

\subsection{Computation complexity analysis}\label{section:complexity_analysis}
The rectification methods often come close in term of image matching accuracy. Another important aspect is their speed. From all three best performing methods (i.e. all affine rectifying methods: DenseAffNet, DepthAffNet and simple AffNetShapes), DenseAffNet was observed to be the fastest, DepthAffNet a bit slower and the simple AffNetShapes rectification is the slowest method. Figure~\ref{fig:running_time} shows the average running times per image rectification via different methods run on the same hardware. It is to provide a general idea about real relative speed. However, as comparing just the running time can sometimes be tricky (depending on various implementation details for example), another perspective is given by statistics in Table~\ref{table:complexity_stats}. Most importantly, the average total area of the rectification warps of the image was observed to be by far the lowest by DenseAffNet and by far the highest by simple AffNetShapes rectification. To summarize, both DepthAffNet and DenseAffNet are faster than the existing methods, while DenseAffNet is the fasters between these two.
\section{Conclusion}
Two methods for image matching with rectification were introduced, which both match or improve the state-of-the-art in accuracy, depending on the dataset. Moreover, they are computationally less intensive. DenseAffNet is faster from the two and achieves better accuracy in more generic IMC Phototourism dataset. Moreover, it is somehow more straightforward as it doesn't require a depth map estimation and can be thus performed with only one CNN. On the other hand, DepthAffNet usage of depth maps seems quite natural and as shown the computation of depth maps does not make the method much slower. Also DepthAffNet seem to be the method of choice for scenes where large planes are present (e.g. the Strong ViewPoint Changes Dataset). 
\section*{Acknowledgements}
Authors are supported by OP VVV funded project CZ.02.1.01/0.0/0.0/$16\_019$/0000765 ``Research Center for Informatics''.

\newpage
\bibliographystyle{plain}
\bibliography{references}
\clearpage

\if\newarxivtemplate1
    \appendix
    \section{Supplementary material}
    \begin{table*}[htb]
\caption{\textbf{Average accuracy of estimated camera pose on St. Peter's scene of IMC Phototourism dataset.}}
\begin{tabular}{crccccc}
\toprule
 min. covisibility & \#pairs & DenseAffNet & DepthAffNet & depth-map-based~\cite{toft} & AffNetShapes~\cite{CnnAssistedCoverings} & unrectified HardNet \\ \midrule
0.2 & 263 & 0.39 & 0.38 & 0.36 & 0.32 & 0.37 \\ 
0.3 & 1322 & 0.66 & 0.66 & 0.66 & 0.66 & 0.67 \\ 
0.4 & 1466 & 0.79 & 0.78 & 0.78 & 0.76 & 0.79 \\ 
0.5 & 1060 & 0.88 & 0.87 & 0.87 & 0.87 & 0.87 \\ 
0.6 & 549 & 0.94 & 0.94 & 0.92 & 0.93 & 0.93 \\ 
0.7 & 224 & 0.93 & 0.92 & 0.92 & 0.92 & 0.93 \\ 
0.8 & 59 & 0.93 & 0.97 & 0.93 & 0.93 & 0.97 \\ 
0.9 & 7 & 1.00 & 1.00 & 1.00 & 1.00 & 1.00 \\ 

\bottomrule
\end{tabular}
\end{table*}

\begin{table*}[htb]
\caption{\textbf{Average accuracy of estimated camera pose on Sacre Coeur scene of IMC Phototourism dataset.}}
\begin{tabular}{crccccc}
\toprule
 min. covisibility & \#pairs & DenseAffNet & DepthAffNet & depth-map-based~\cite{toft} & AffNetShapes~\cite{CnnAssistedCoverings} & unrectified HardNet \\ \midrule
0.1 & 484 & 0.56 & 0.57 & 0.56 & 0.55 & 0.70 \\ 
0.2 & 1093 & 0.86 & 0.85 & 0.83 & 0.86 & 0.89 \\ 
0.3 & 1464 & 0.94 & 0.91 & 0.89 & 0.94 & 0.94 \\ 
0.4 & 1069 & 0.95 & 0.92 & 0.90 & 0.95 & 0.95 \\ 
0.5 & 504 & 0.96 & 0.95 & 0.93 & 0.96 & 0.95 \\ 
0.6 & 204 & 0.98 & 0.97 & 0.94 & 0.97 & 0.98 \\ 
0.7 & 103 & 0.97 & 0.98 & 0.92 & 0.98 & 0.97 \\ 
0.8 & 24 & 1.00 & 1.00 & 0.96 & 1.00 & 1.00 \\ 
0.9 & 5 & 1.00 & 1.00 & 1.00 & 1.00 & 1.00 \\ 
\bottomrule
\end{tabular}
\end{table*}

\begin{table*}[htb]
\caption{\textbf{Average accuracy of estimated camera pose on Reichstag scene of IMC Phototourism dataset.}}
\begin{tabular}{crccccc}
\toprule
 min. covisibility & \#pairs & DenseAffNet & DepthAffNet & depth-map-based~\cite{toft} & AffNetShapes~\cite{CnnAssistedCoverings} & unrectified HardNet \\ \midrule
0.2 & 147 & 0.64 & 0.67 & 0.63 & 0.58 & 0.64 \\ 
0.3 & 763 & 0.88 & 0.85 & 0.84 & 0.88 & 0.89 \\ 
0.4 & 680 & 0.96 & 0.90 & 0.90 & 0.97 & 0.97 \\ 
0.5 & 869 & 0.99 & 0.99 & 0.98 & 0.99 & 0.99 \\ 
0.6 & 192 & 0.98 & 0.98 & 0.99 & 0.99 & 0.98 \\ 
0.7 & 23 & 0.96 & 1.00 & 0.96 & 0.96 & 0.96 \\ 
0.8 & 18 & 1.00 & 1.00 & 1.00 & 1.00 & 1.00 \\ 
0.9 & 9 & 1.00 & 1.00 & 1.00 & 1.00 & 1.00 \\ 

\bottomrule
\end{tabular}
\end{table*}   

\begin{table*}[htb]
\caption{\textbf{Average accuracy of estimated camera pose on scene 1 of Extreme View Dataset.}}
\begin{tabular}{rrccccc}
\toprule
 min. rotation & \#pairs & DenseAffNet & DepthAffNet & depth-map-based~\cite{toft} & AffNetShapes~\cite{CnnAssistedCoverings} & unrectified HardNet \\ \midrule
0$^{\circ}$ & 200 & 0.95 & 0.94 & 0.92 & 0.94 & 0.93 \\ 
10$^{\circ}$ & 200 & 0.92 & 0.94 & 0.90 & 0.91 & 0.90 \\ 
20$^{\circ}$ & 200 & 0.89 & 0.91 & 0.89 & 0.89 & 0.88 \\ 
30$^{\circ}$ & 200 & 0.87 & 0.92 & 0.82 & 0.90 & 0.87 \\ 
40$^{\circ}$ & 200 & 0.85 & 0.88 & 0.77 & 0.90 & 0.82 \\ 
50$^{\circ}$ & 200 & 0.68 & 0.74 & 0.70 & 0.76 & 0.63 \\ 
60$^{\circ}$ & 200 & 0.58 & 0.62 & 0.58 & 0.64 & 0.58 \\ 
70$^{\circ}$ & 200 & 0.56 & 0.64 & 0.61 & 0.65 & 0.52 \\ 
80$^{\circ}$ & 200 & 0.54 & 0.56 & 0.55 & 0.60 & 0.45 \\ 
90$^{\circ}$ & 200 & 0.50 & 0.55 & 0.52 & 0.58 & 0.45 \\ 
100$^{\circ}$ & 200 & 0.50 & 0.61 & 0.48 & 0.58 & 0.44 \\ 
110$^{\circ}$ & 200 & 0.42 & 0.52 & 0.40 & 0.51 & 0.30 \\ 
120$^{\circ}$ & 200 & 0.39 & 0.52 & 0.39 & 0.48 & 0.18 \\ 
130$^{\circ}$ & 200 & 0.37 & 0.44 & 0.35 & 0.42 & 0.11 \\ 
140$^{\circ}$ & 200 & 0.14 & 0.19 & 0.08 & 0.13 & 0.01 \\ 
150$^{\circ}$ & 200 & 0.08 & 0.12 & 0.02 & 0.07 & 0.04 \\ 
160$^{\circ}$ & 200 & 0.01 & 0.04 & 0.00 & 0.02 & 0.00 \\ 
170$^{\circ}$ & 190 & 0.00 & 0.02 & 0.00 & 0.00 & 0.00 \\ 

\bottomrule
\end{tabular}
\end{table*}

\begin{table*}[htb]
\caption{\textbf{Average accuracy of estimated camera pose on scenes 2 to 8 of Extreme View Dataset.}}
\begin{tabular}{rrccccc}
\toprule
 min. rotation & \#pairs & DenseAffNet & DepthffNet & depth-map-based~\cite{toft} & AffNetShapes~\cite{CnnAssistedCoverings} & unrectified HardNet \\ \midrule
0$^{\circ}$ & 1400 & 0.94 & 0.96 & 0.93 & 0.95 & 0.94 \\ 
10$^{\circ}$ & 1400 & 0.86 & 0.86 & 0.82 & 0.86 & 0.85 \\ 
20$^{\circ}$ & 1400 & 0.78 & 0.79 & 0.72 & 0.81 & 0.77 \\ 
30$^{\circ}$ & 1400 & 0.72 & 0.71 & 0.62 & 0.73 & 0.70 \\ 
40$^{\circ}$ & 1400 & 0.63 & 0.63 & 0.55 & 0.65 & 0.61 \\ 
50$^{\circ}$ & 1400 & 0.57 & 0.55 & 0.48 & 0.58 & 0.53 \\ 
60$^{\circ}$ & 1400 & 0.47 & 0.47 & 0.42 & 0.50 & 0.46 \\ 
70$^{\circ}$ & 1400 & 0.44 & 0.42 & 0.36 & 0.45 & 0.40 \\ 
80$^{\circ}$ & 1400 & 0.39 & 0.39 & 0.31 & 0.41 & 0.35 \\ 
90$^{\circ}$ & 1400 & 0.34 & 0.34 & 0.25 & 0.36 & 0.32 \\ 
100$^{\circ}$ & 1400 & 0.30 & 0.32 & 0.20 & 0.32 & 0.26 \\ 
110$^{\circ}$ & 1400 & 0.28 & 0.30 & 0.18 & 0.32 & 0.22 \\ 
120$^{\circ}$ & 1400 & 0.23 & 0.28 & 0.13 & 0.29 & 0.16 \\ 
130$^{\circ}$ & 1212 & 0.12 & 0.14 & 0.05 & 0.16 & 0.09 \\ 
140$^{\circ}$ & 997 & 0.12 & 0.13 & 0.05 & 0.15 & 0.10 \\ 
150$^{\circ}$ & 881 & 0.10 & 0.09 & 0.03 & 0.09 & 0.07 \\ 
160$^{\circ}$ & 746 & 0.05 & 0.05 & 0.03 & 0.05 & 0.04 \\ 
170$^{\circ}$ & 440 & 0.02 & 0.01 & 0.01 & 0.02 & 0.01 \\ 

\bottomrule
\end{tabular}

\end{table*}

\fi

\end{document}